\documentclass{article}

\usepackage{microtype}
\usepackage{graphicx}
\usepackage{subfigure}
\usepackage{booktabs} 

\usepackage{hyperref}

\usepackage[accepted]{icml2024}

\usepackage{amsmath}
\usepackage{amssymb}
\usepackage{mathtools}
\usepackage{amsthm}
\usepackage{arydshln}
\usepackage[capitalize,noabbrev]{cleveref}

\theoremstyle{plain}
\newtheorem{theorem}{Theorem}[section]

\theoremstyle{definition}
\newtheorem{definition}[theorem]{Definition}

\theoremstyle{remark}

\usepackage{multirow}
\usepackage{diagbox}

\newcommand{\vpara}[1]{\vspace{0.05in}\noindent\textbf{#1 }}

\usepackage{xcolor}

\usepackage[textsize=tiny]{todonotes}

\icmltitlerunning{Exploring Correlations of Self-Supervised Tasks for Graphs}

\begin{document}

\twocolumn[
\icmltitle{Exploring Correlations of Self-Supervised Tasks for Graphs}




\begin{icmlauthorlist}
\icmlauthor{Taoran Fang}{zju}
\icmlauthor{Wei Zhou}{zju}
\icmlauthor{Yifei Sun}{zju}
\icmlauthor{Kaiqiao Han}{zju}
\icmlauthor{Lvbin Ma}{huayun}
\icmlauthor{Yang Yang}{zju}
\end{icmlauthorlist}

\icmlaffiliation{zju}{College of Computer Science and Technology, Zhejiang University, Hangzhou, China}
\icmlaffiliation{huayun}{Zhejiang Huayun Information Technology, Hangzhou, China. Our code is available at: \url{https://github.com/zjunet/GraphTCM}}

\icmlcorrespondingauthor{Yang Yang}{yangya@zju.edu.cn}

\icmlkeywords{Machine Learning, ICML}

\vskip 0.3in
]



\printAffiliationsAndNotice{}  

\begin{abstract}

Graph self-supervised learning has sparked a research surge in training informative representations without accessing any labeled data.
However, our understanding of graph self-supervised learning remains limited, and the inherent relationships between various self-supervised tasks are still unexplored.
Our paper aims to provide a fresh understanding of graph self-supervised learning based on task correlations.
Specifically, we evaluate the performance of the representations trained by one specific task on other tasks and define correlation values to quantify task correlations.
Through this process, we unveil the task correlations between various self-supervised tasks and can measure their expressive capabilities, which are closely related to downstream performance.
By analyzing the correlation values between tasks across various datasets, we reveal the complexity of task correlations and the limitations of existing multi-task learning methods.
To obtain more capable representations, we propose Graph Task Correlation Modeling (GraphTCM) to illustrate the task correlations and utilize it to enhance graph self-supervised training.
The experimental results indicate that our method significantly outperforms existing methods across various downstream tasks. 

\end{abstract}

\section{Introduction}

Graph neural networks (GNNs) have garnered widespread attention from researchers due to their remarkable success in graph representation learning~\citep{kipf2016semi,hamilton2017inductive,Xu2019HowPA,Wu2019ACS}. 
Most existing studies on GNNs have focused on (semi-)supervised scenarios and have been widely applied in various downstream tasks such as node classification \citep{Jiang2019SemiSupervisedLW}, link prediction \citep{kipf2016variational}, vertex clustering \citep{ramaswamy2005distributed}, and recommendation systems \citep{ying2018graph}.
However, two fundamental challenges hinder the large-scale practical applications of GNNs.
One is the scarcity of high-quality labeled data in the real world~\citep{Zitnik2018PrioritizingNC}, and the other is the low generalization ability of the trained models caused by over-fitting~\citep{rong2019dropedge,feng2020graph,Fang2022DropMessageUR}.

To address these challenges, graph self-supervised learning~\citep{Liu2021GraphSL,Wu2021SelfSupervisedLO,Xie2021SelfSupervisedLO} offers a promising training paradigm to reduce the reliance on task labels and demonstrates significant potential in strengthening GNNs in an unsupervised manner.
These methods specifically construct a variety of graph self-supervised tasks based on structural and attribute information from the training graphs, wherein the supervision signals are automatically acquired from the graph itself without accessing any labeled data.
By employing well-designed graph self-supervised tasks, these methods enable the GNN model to extract more informative representations from unlabeled data, leading to improved generalization~\citep{Hu2020GPTGNNGP,Qiu2020GCCGC,Hu2020StrategiesFP} and robustness~\citep{You2020WhenDS,Jovanovic2021TowardsRG} on downstream datasets.

However, two fundamental problems impede further comprehension of graph self-supervised learning.
The first problem is the lack of understanding regarding the correlations between different graph self-supervised tasks.
Despite the multitude of graph self-supervised tasks with diverse training objectives, discussions on the correlations between these tasks are scarce.
Additionally, the differences in trained representations between various self-supervised tasks remain unexplored, limiting a comprehensive understanding of the expressive capabilities of the trained representations.
The second problem revolves around how to train sufficiently general and robust self-supervised representations.
While existing graph self-supervised methods can train representations that excel on their respective tasks, there is no direct guarantee of the performance of trained representations on downstream tasks. 
In practice, downstream tasks encompass diverse forms, posing a significant challenge in ensuring the universality of representations trained through self-supervised tasks.

In this paper, we present a novel perspective on graph self-supervised learning based on trained representations.
Specifically, we obtain trained representations of these tasks on various datasets and assess the performance of these trained representations on other tasks.
To address the first problem, we introduce \textit{Correlation Values} to measure the correlation between two tasks quantitatively.
More precisely, the correlation value $\text{Cor}(t_1,t_2)$ is defined as the relative loss when using representations trained by task $t_1$ on the objective function of task $t_2$.
A lower value of $\text{Cor}(t_1,t_2)$ indicates that the representations trained by task $t_1$ perform better on task $t_2$.
Furthermore, we propose Average Task Difficulty (ATD) and Average Relative Loss (ARL) to compare the task difficulty and the representation capability between various tasks, providing a comprehensive understanding of the expressive capabilities of the representations trained by this self-supervised task. 

To tackle the second problem, we theoretically and empirically demonstrate that if trained representations excel across various self-supervised tasks, they are also likely to exhibit universality and effectiveness across a range of downstream tasks.
In pursuing this objective, several pioneering works~\citep{Jin2021AutomatedSL,Ju2022MultitaskSG,Han2021AdaptiveTL} apply multi-task learning to combine the losses of different self-supervised tasks adaptively.
However, upon evaluating the performance of the representations trained by these methods across various self-supervised tasks, our results indicate that these methods are unable to train representations that perform well across all self-supervised tasks, and they are even unable to surpass the representations trained by a single task.
From a completely new perspective, we introduce GraphTCM to model the task correlations among various graph self-supervised tasks and utilize it to obtain representations that perform effectively on all involved self-supervised tasks.
With a well-trained GraphTCM, we can obtain trained representations that perform competently across various self-supervised and downstream tasks by only introducing a linear transformation with limited parameters.
Overall, the contributions of our work can be summarized as follows:
\begin{itemize}
    \item {Providing a new understanding of graph self-supervised learning based on trained representations, we calculate the correlation values to obtain quantitative characterizations of the correlations between various graph self-supervised tasks and the expressive capabilities of their trained representations.  
    }
    \item{We explore the intricate correlations of various graph self-supervised tasks across diverse real datasets and verify that existing multi-task learning methods fail to train representations that perform effectively across all self-supervised tasks.}
    \item{We introduce GraphTCM to model task correlations and utilize it to enhance self-supervised training. Experimental results demonstrate that our GraphTCM-enhanced representations not only exhibit strong performance across all self-supervised tasks but also significantly outperform existing methods across all downstream tasks.}
\end{itemize}

\section{Related work}

\vpara{Graph Self-supervised Learning.}
Graph self-supervised learning addresses the scarcity of labeled data and combats low generalization in graph models by reducing reliance on task-specific labels and designing tasks capable of mining knowledge of input graphs~\citep{Wu2021SelfSupervisedLO,Xie2021SelfSupervisedLO}.
Existing graph self-supervised tasks can be primarily categorized into four types: feature-based, structure-based, auxiliary property-based, and contrastive-based tasks~\citep{Liu2021GraphSL}.
Specifically, feature-based tasks~\citep{Jin2020SelfsupervisedLO,You2020WhenDS,hou2022graphmae,Hu2020StrategiesFP,Wang2017MGAEMG,Manessi2020GraphBasedNN,Park2019SymmetricGC} focus on reconstructing the graph feature, while structure-based tasks~\citep{kipf2016variational,li2023seegera,Hasanzadeh2019SemiImplicitGV,Pan2018AdversariallyRG,Kim2022HowTF,Hu2019PreTrainingGN} reconstruct the graph adjacency or masked edges.
Meanwhile, auxiliary property-based tasks~\citep{Jin2020SelfsupervisedLO,You2020WhenDS,Sun2019MultiStageSL,Zhu2020CAGNNCG,Peng2020SelfSupervisedGR} guide models by predicting certain pre-computable auxiliary information (\emph{e.g.}, node degree).
Contrast-based tasks are also widely used in practice.
These tasks~\citep{Velickovic2019DeepGI,jiao2020sub,You2020GraphCL,Qiu2020GCCGC,xu2022ccgl} do not explicitly incorporate the desired information as the prediction targets, but instead, they focus on the mutual information (MI) maximization~\citep{hjelm2019learning} where the estimated MI between augmented instances of the same object is maximized.

\vpara{Graph Multi-task Learning.}
Several pioneering approaches have employed multi-task training methods to integrate various graph self-supervised tasks to acquire representations that can effectively perform across a spectrum of self-supervised and downstream tasks.
AUX-TS~\citep{Han2021AdaptiveTL} introduces an adaptive mechanism for selecting and combining diverse pre-training tasks with target tasks during the fine-tuning phase.
AutoSSL~\citep{Jin2021AutomatedSL} integrates multiple self-supervised tasks using learnable coefficients, leveraging the homophily assumption as a basis for the integration process.
ParetoGNN~\citep{Ju2022MultitaskSG} aims to dynamically adjust the coefficients of task combination during the self-supervised training process, with the objective of facilitating Pareto optimality.
However, these methods primarily concentrate on combining losses from different tasks, thus neglecting the disparity between the trained and ideal representations and failing to characterize the trained representations adequately.

\section{Exploring the Task Correlations}
\label{sec:method}

\subsection{Preliminaries}

\vpara{Notations.}
Let $\mathcal{G}=\left(\mathcal{V}, \mathcal{E}\right) \in \mathbb{G}$ represents a graph, 
where $\mathcal{V}=\left\{v_{1}, v_{2}, \ldots, v_{N}\right\}$, $\mathcal{E} \subseteq \mathcal{V} \times \mathcal{V}$ denote the node set and edge set respectively.
The node features can be denoted as a matrix 
$\mathbf{X} = \left\{x_{1}, x_{2}, \ldots, x_{N} \right\} \in \mathbb{R}^{N \times F}$, 
where $x_{i}\in\mathbb{R}^{F}$ is the feature of the node $v_i$, and $F$ is the dimensionality of node features.
$\mathbf{A}\in \{0,1\}^{N\times N}$ denotes the adjacency matrix, where $\mathbf{A}_{ij}=1$ if $(v_i,v_j)\in \mathcal{E}$.

\vpara{Graph Self-supervised Learning.}
For a GNN model $g$ parameterized by $\theta$, a linear tuning head $\mathbf{W}$ (possibly not required), a graph $\mathcal{G}:(\mathbf{A},\mathbf{X})$ and a self-supervised target $\mathbf{Y}_{\text{ssl}}$, we search for the parameters $\hat{\theta}$ and $\hat{\mathbf{W}}$ that minimize the self-supervised loss $\boldsymbol{l}_{\text{ssl}}$ during the training process, which can be expressed as:
\begin{align}
\hat{\theta}, \hat{\mathbf{W}}=\mathop{\arg\min}_{\theta,\mathbf{W}} \ \boldsymbol{l}_{\text{ssl}}(g(\mathbf{A,X})\cdot\mathbf{W},\mathbf{Y}_{\text{ssl}})
\label{formula:ssl}
\end{align}
Existing graph self-supervised methods can be categorized into four primary types~\citep{Liu2021GraphSL}: \textit{feature-based (FB)}, \textit{structure-based (SB)}, \textit{auxiliary property-based (APB)} and \textit{contrast-based (CB)}.
To comprehensively understand the complex relationships in graph self-supervised tasks, we have chosen two representative methods from each category for detailed analysis.
Table \ref{tab:methods} provides a summary of their training targets, and their detailed descriptions can be found in Appendix \ref{appendix:detailed_method}.

\begin{table}[h]
    \vspace{-0.3cm}
    \centering
    \caption{Overview of the involved self-supervised tasks.}
    \label{tab:methods}
    \resizebox{1\columnwidth}{!}{
    \begin{tabular}{ccc}
    \toprule
    Method & Category & Training Target \\
    \midrule
    GraphComp~\citep{You2020WhenDS} & FB & Node Feature \\
    AttributeMask~\citep{Jin2020SelfsupervisedLO} & FB & PCA Node Feature \\
    GAE~\citep{Kipf2016VariationalGA} & SB & Adjacency Matrix \\
    EdgeMask~\citep{Jin2020SelfsupervisedLO} & SB & Masked Edge \\
    NodeProp~\citep{Jin2020SelfsupervisedLO} & APB & Degree Calculation \\
    DisCluster~\citep{Jin2020SelfsupervisedLO} & APB & Distance Calculation \\
    DGI~\citep{Velickovic2018DeepGI} & CB & Cross-scale Contrast \\
    SubgCon~\citep{jiao2020sub} & CB & Same-scale Contrast \\
    \bottomrule
    \end{tabular}}
    \vspace{-0.3cm}
\end{table}

\begin{figure*}[t]
\centering
\includegraphics[width=\textwidth]{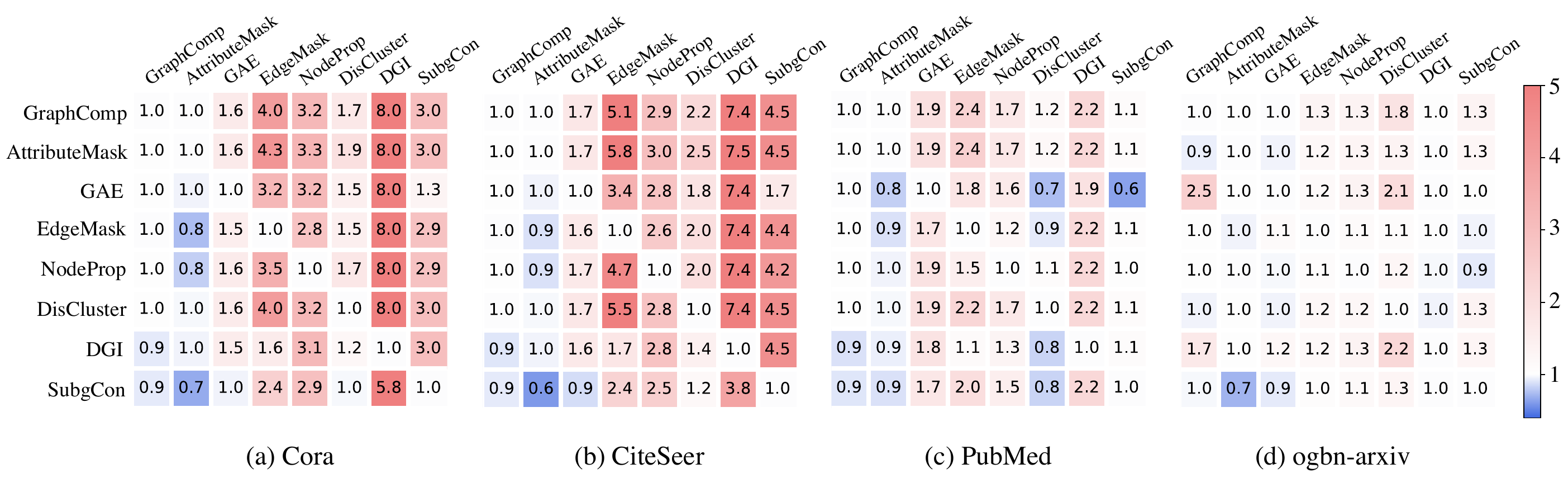}
\vspace{-0.5cm}
\caption{The real correlation values on various graph datasets. Here, the $y$-axis represents the task used to train the assessed representations, while the $x$-axis represents the self-supervised task on which the representations are evaluated. For instance, the intersection of the DGI row and the GAE column is denoted as $\text{Cor}(\text{DGI}, \text{GAE})$ according to Formula \ref{formula:cor}, signifying the comparative performance of the representations trained by DGI on the GAE task. A lower value indicates better performance.}
\label{fig:heatmap}
\end{figure*}

\subsection{Building the Task Correlations}

We transfer trained representations from one self-supervised task to other tasks and evaluate the correlation between self-supervised tasks by examining the performance of the transferred representations on these tasks.
According to Formula \ref{formula:ssl}, the training process of graph self-supervision can be interpreted as an optimization process for the self-supervised objective function.
Consequently, assessing the loss of representations for a self-supervised task can indicate whether the representation possesses the expressive capability required for that particular task.
In contrast to subjective judgments based on training objectives or processes for different self-supervised tasks, our assessment is objective and definitive, revealing the actual characteristics of various self-supervised methods in practical scenarios.

For a graph $\mathcal{G}:(\mathbf{A},\mathbf{X})$, we denote the set of self-superivsed tasks as $\mathcal{T}=\{t_1,t_2,...,t_k\}$, where $t_i$ denotes a specific self-supervised task.
Additionally, we denote the loss function of the task $t_i$ as $\boldsymbol{l}_{t_i}(\cdot)$, and denote the target matrix of the task $t_i$ as $\mathbf{Y}_{t_i}$.
Considering a GNN model $g$ and a possible linear tuning head $\mathbf{W}$, we assume the model obtained through gradient descent based on the loss function $\boldsymbol{l}_{t_i}(g(\mathbf{A},\mathbf{X})\cdot \mathbf{W},\mathbf{Y}_{t_i})$ is denoted as $g_{t_i}$.
Furthermore, we denote the graph representations obtained by $g_{t_i}$ as $\mathbf{H}_{t_i}=g_{t_i}(\mathbf{A},\mathbf{X})$.

\begin{definition}
\label{def:cor_value}
\textit{(Correlation Value)} Given two self-supervised tasks $t_1,t_2\in \mathcal{T}$, a graph $\mathcal{G}:(\mathbf{A},\mathbf{X})$, we define the correlation value $\text{Cor}(t_1,t_2)$ as:
\begin{align}
\label{formula:cor}
\text{Cor}(t_1,t_2)=\frac{\min_{\mathbf{W}_{t_1}}\boldsymbol{l}_{t_2}(\mathbf{H}_{t_1}\cdot\mathbf{W}_{t_1},\mathbf{Y}_{t_2})}{\min_{\mathbf{W}_{t_2}}\boldsymbol{l}_{t_2}(\mathbf{H}_{t_2}\cdot\mathbf{W}_{t_2},\mathbf{Y}_{t_2})}
\end{align}
\end{definition}

Both the numerator and the denominator are optimized using gradient descent.
It is important to note that the linear tuning heads $\mathbf{W}_{t_1}$ and $\mathbf{W}_{t_2}$ in the numerator and denominator of $\text{Cor}(t_1,t_2)$ do not have to be identical.
$\text{Cor}(t_1,t_2)$ reflects the relative performance of the representations $\mathbf{H}_{t_1}$ trained by task $t_1$ on the task $t_2$.
It is evident that $\text{Cor}(t_i,t_i)$ is equal to $1$.
The magnitude of the correlation values can be elucidated from two perspectives: \textit{representation capability} and \textit{task difficulty}.
On one hand, $\text{Cor}(t_1,t_2)$ indicates the capability of the representations $\mathbf{H}_{t_1}$. 
If $\text{Cor}(t_1,t_2)$ is close to or less than 1, it signifies that the representations trained by task $t_1$ possess the expressive capability required for task $t_2$.
On the other hand, $\text{Cor}(t_1,t_2)$ also indicates the difficulty of $t_2$.
If $\text{Cor}(t_1,t_2)$ is large, it implies that task $t_2$ is challenging for the representations trained by task $t_1$.
To further delineate representation capability and task difficulty, we introduce the following definitions.

\begin{definition}
\textit{(Average Task Difficulty)} Given a graph $\mathcal{G}:(\mathbf{A},\mathbf{X})$, a set of self-superivsed tasks $\mathcal{T}=\{t_1,t_2,...,t_k\}$, the average task difficulty (ATD) for task $t_i$ is calculated as:
\begin{align}
\text{ATD}_{t_i}=\frac{1}{k-1}\sum_{j\neq i}^{k}\text{Cor}(t_j,t_i)
\end{align}
\end{definition}

\begin{definition}
\textit{(Average Relative Loss)} Given a graph $\mathcal{G}:(\mathbf{A},\mathbf{X})$, a set of self-superivsed tasks $\mathcal{T}=\{t_1,t_2,...,t_k\}$, the average relative loss (ARL) for task $t_i$ is calculated as:
\begin{align}
\text{ARL}_{t_i}=\frac{1}{k}(\text{ATD}_{t_i}+\sum_{j\neq i}^{k}\text{Cor}(t_i,t_j))
\end{align}
\end{definition}

ATD is directly associated with task difficulty, indicating that the more challenging a task is, the higher the ATD value it achieves. 
As for ARL, the scenario is different. 
A low ARL value for a task signifies that the representations trained by this task exhibit high representation capability, as they perform well on other tasks.
When computing $\text{ARL}_{t_i}$, we incorporate an additional $\text{ATD}_{t_i}$ before calculating the average to prevent a task from disproportionately benefiting due to its high task difficulty.
This adjustment enables a more equitable comparison of representation capabilities across different tasks.
Subsequently, we will illustrate the intricate correlations among various graph self-supervised tasks across diverse datasets.

\subsection{Unveiling the T ask Correlations}

In this chapter, we compute the correlation values for the 8 graph self-supervised tasks outlined in Table 1 across 4 graph datasets (\emph{i.e.}, Cora, CiteSeer, PubMed and ogbn-arxiv).
A concise description of these datasets can be found in Appendix \ref{appendix:detail_dataset}.
Figure \ref{fig:heatmap} illustrates the correlation values among various tasks across these four datasets.
Additionally, Table \ref{tab:ATD} and \ref{tab:ARL} summarize the values and rankings of ATD and ARL, respectively.

\begin{table}[htbp]
    \centering
    \vspace{-0.3cm}
    \caption{The statistics of ATD for various self-supervised tasks (the larger value indicates higher task difficulty $\uparrow$). The numbers in the parentheses signify the \textit{ranking} of task ATD within the same dataset, where a smaller number implies greater task difficulty.}
    \label{tab:ATD}
    \resizebox{1\columnwidth}{!}{
    \begin{tabular}{c|cccc}
    \toprule
    Task & Cora & CiteSeer & PubMed & ogbn-arxiv \\
    \midrule
    GraphComp	& 1.00 ($\mathbf{7}$)	& 0.99 ($\mathbf{7}$)	& 0.99 ($\mathbf{6}$)	& 1.30 ($\mathbf{2}$) \\
    AttributeMask	& 0.88 ($\mathbf{8}$)	& 0.90 ($\mathbf{8}$)	& 0.93 ($\mathbf{8}$)	& 0.97 ($\mathbf{8}$) \\
    GAE	& 1.48 ($\mathbf{6}$)	& 1.56 ($\mathbf{6}$)	& 1.85 ($\mathbf{3}$)	& 1.02 ($\mathbf{6}$) \\
    EdgeMask	& 3.29 ($\mathbf{2}$)	& 4.08 ($\mathbf{2}$)	& 1.92 ($\mathbf{2}$)	& 1.17 ($\mathbf{4}$) \\
    NodeProp	& 3.10 ($\mathbf{3}$)	& 2.79 ($\mathbf{4}$)	& 1.53 ($\mathbf{4}$)	& 1.25 ($\mathbf{3}$) \\
    DisCluster	& 1.49 ($\mathbf{5}$)	& 1.87 ($\mathbf{5}$)	& 0.96 ($\mathbf{7}$)	& 1.57 ($\mathbf{1}$) \\
    DGI	& 7.69 ($\mathbf{1}$)	& 6.92 ($\mathbf{1}$)	& 2.13 ($\mathbf{1}$)	& 1.00 ($\mathbf{7}$) \\
    SubgCon	& 2.73 ($\mathbf{4}$)	& 4.04 ($\mathbf{3}$)	& 1.01 ($\mathbf{5}$)	& 1.16 ($\mathbf{5}$) \\
    \bottomrule
    \end{tabular}
    }
    \vspace{-0.3cm}
\end{table}

\begin{table}[htbp]
    \centering
    \vspace{-0.3cm}
    \caption{The statistics of ARL for various self-supervised tasks (the lower value indicates stronger representation capability $\downarrow$). The numbers in the parentheses signify the \textit{ranking} of task ARL within the same dataset, where a smaller number implies a stronger representation capability.}
    \label{tab:ARL}
    \resizebox{1\columnwidth}{!}{
    \begin{tabular}{c|cccc}
    \toprule
    Task & Cora & CiteSeer & PubMed & ogbn-arxiv \\
    \midrule
    GraphComp	& 2.93 ($\mathbf{7}$)	& 3.23 ($\mathbf{7}$)	& 1.55 ($\mathbf{7}$)	& 1.27 ($\mathbf{6}$) \\
    AttributeMask	& 3.00 ($\mathbf{8}$)	& 3.36 ($\mathbf{8}$)	& 1.56 ($\mathbf{8}$)	& 1.12 ($\mathbf{4}$) \\
    GAE	& 2.58 ($\mathbf{3}$)	& 2.59 ($\mathbf{2}$)	& 1.29 ($\mathbf{2}$)	& 1.39 ($\mathbf{8}$) \\
    EdgeMask	& 2.71 ($\mathbf{4}$)	& 3.00 ($\mathbf{4}$)	& 1.37 ($\mathbf{3}$)	& 1.06 ($\mathbf{2}$) \\
    NodeProp	& 2.82 ($\mathbf{5}$)	& 3.09 ($\mathbf{5}$)	& 1.40 ($\mathbf{5}$)	& 1.06 ($\mathbf{3}$) \\
    DisCluster	& 2.91 ($\mathbf{6}$)	& 3.23 ($\mathbf{6}$)	& 1.51 ($\mathbf{6}$)	& 1.15 ($\mathbf{5}$) \\
    DGI	& 2.51 ($\mathbf{2}$)	& 2.62 ($\mathbf{3}$)	& 1.27 ($\mathbf{1}$)	& 1.36 ($\mathbf{7}$) \\
    SubgCon	& 2.18 ($\mathbf{1}$)	& 2.03 ($\mathbf{1}$)	& 1.38 ($\mathbf{4}$)	& 1.03 ($\mathbf{1}$) \\
    \bottomrule
    \end{tabular}
    }
\end{table}

\vpara{Overall Analysis.}
The correlation values depict the inherent relationships among different graph self-supervised tasks.
According to our findings, the relationships between graph self-supervised tasks are considerably more intricate than previously assumed.

\textit{The correlations among graph self-supervised tasks exhibit variability across different datasets.}
In contrast to the consistent correlations observed for self-supervised tasks in other domains~\citep{Zamir2018TaskonomyDT}, the correlations among graph self-supervised tasks fluctuate with shifts in the dataset, suggesting that merely considering the correlations of graph self-supervised tasks from the standpoint of training processes or objective functions is insufficient.

\textit{A straightforward relationship between representation capability and task difficulty is absent.}
The tables in \ref{tab:ATD} and \ref{tab:ARL} indicate that task difficulty and representation capability do not consistently display a positive or negative correlation.
This suggests that opting for more challenging tasks does not automatically improve representation capability.


\textit{Feature-based graph self-supervised tasks generally exhibit lower task difficulty and are often overlapped by other tasks.}
This phenomenon may arise from the utilization of node features as auxiliary information in other self-supervised tasks, leading to representations learned by these tasks also encompassing information about node features.


\vpara{The performance of multi-task training.}
To obtain representations that perform effectively across various tasks, most existing methods explore solutions from the perspective of multi-task training.
Thus, \textit{can we attain more capable representations through multi-task training?}
To answer this question, we assess the ARL of representations obtained through three multi-task training methods (Multi-loss, AutoSSL~\citep{Jin2021AutomatedSL}, ParetoGNN~\citep{Ju2022MultitaskSG}) as well as two straightforward representation mixing methods (representation addition and representation concat).
The specific implementations for these methods can be found in Appendix \ref{appendix:detail_multi_task}, with the base tasks for their multi-task training involving eight self-supervised tasks.
Table \ref{tab:ARL_bl} presents a summary of the evaluation results.
\begin{table}[htbp]
    \centering
    \vspace{-0.3cm}
    \caption{The statistics of ARL for various multi-task representations. The numbers within the parentheses signify the \textit{ranking} of the ARL values compared with the results in Table \ref{tab:ARL}, where a smaller number implies stronger representations capability.}
    \label{tab:ARL_bl}
    \resizebox{1\columnwidth}{!}{
    \begin{tabular}{c|cccc}
    \toprule
    Method & Cora & CiteSeer & PubMed & ogbn-arxiv \\
    \midrule
    Addition	& 2.71 ($\mathbf{5}$)	& 2.79 ($\mathbf{4}$)	& 2.17 ($\mathbf{9}$)	& 2.78 ($\mathbf{9}$) \\
    Concat	& 2.32 ($\mathbf{2}$)	& 2.26 ($\mathbf{2}$)	& 1.32 ($\mathbf{3}$)	& 1.82 ($\mathbf{9}$) \\
    Multi-loss	& 2.81 ($\mathbf{6}$)	& 3.32 ($\mathbf{8}$)	& 1.51 ($\mathbf{7}$)	& 1.92 ($\mathbf{9}$) \\
    AutoSSL	& 4.53 ($\mathbf{9}$)	& 2.73 ($\mathbf{4}$)	& 2.37 ($\mathbf{9}$)	& 2.38 ($\mathbf{9}$) \\
    ParetoGNN	& 3.01 ($\mathbf{9}$)	& 2.94 ($\mathbf{4}$)	& 2.31 ($\mathbf{9}$)	& 1.90 ($\mathbf{9}$) \\
    \bottomrule
    \end{tabular}
    }
\end{table}
From the findings presented in Table \ref{tab:ARL_bl}, it is evident that multi-task training scarcely enhances representation capability.
These methods do not outperform the representation capability of single tasks on any dataset, and in certain instances, they even demonstrate notable declines in representation capability (\emph{e.g.}, ogbn-arxiv).
This suggests that optimizing the losses of multiple tasks together not only falls short in training representations with versatile capabilities, but also significantly compromises representation capability.
The reasons for this phenomenon might stem from the varying scales of loss for different self-supervised tasks~\citep{Jin2021AutomatedSL}, making their combination challenging; the different optimization directions of various self-supervised tasks leading to confusion when aggregated~\citep{Han2021AdaptiveTL}; and the inability of existing multi-task training methods to capture the intrinsic relationships between individual self-supervised tasks during the training process.

\subsection{Modeling the Task Correlations}

The correlation values among different self-supervised tasks signify the intrinsic relationships between these tasks.
However, existing graph self-supervised methods have not considered these vital relationships.
We are the first to concentrate on the task correlations among graph self-supervised tasks and to leverage these correlations to enhance representation training.
To model the correlations between self-supervised tasks, we introduce the \textbf{Graph} \textbf{T}ask \textbf{C}orrelation \textbf{M}odeling (GraphTCM).
Subsequently, we employ this correlation-auxiliary model to assist us in training representations with high representation capability.

\vpara{Overview.}
Our objective is to characterize the correlation values between different graph self-supervised tasks.
Based on our earlier analysis, correlation values will fluctuate with changes in datasets.
For a certain dataset $\mathcal{D}$, we have:
\begin{align}
\text{Cor}(t_i,t_j)=f_{\mathcal{D}}(t_i,t_j)
\end{align}
where $t_i$ and $t_j$ are two distinct tasks, and $f_{\mathcal{D}}$ is a specific mapping determined by the dataset.

\vpara{Graph Task Correlation Modeling.}
GraphTCM leverages the computed correlation values as ground truth, aiming to characterize the mapping $f_{\mathcal{D}}$.
For a dataset $\mathcal{D}$, we utilize the representation $\mathbf{H}_{t_i}$ and $\mathbf{H}_{t_j}$ trained by tasks $t_i$ and $t_j$ as inputs to derive their correlation value, expressed as:
\begin{align}
\text{GraphTCM}_{\mathcal{D}}(\mathbf{H}_{t_i},\mathbf{H}_{t_j}) \rightarrow \text{Cor}(t_i,t_j)
\end{align}
The concrete implementation of GraphTCM can be expressed as:
\begin{align}
q_{t_i}=\text{Readout}(\mathbf{H}_{t_i}\cdot\mathbf{W}_r) &\quad k_{t_j}=\text{Readout}(\mathbf{H}_{t_j}\cdot\mathbf{W}_t) \nonumber \\
\text{GraphTCM}_{\mathcal{D}}(\mathbf{H}_{t_i}&,\mathbf{H}_{t_j})=\exp({q_{t_i} \cdot k_{t_j}^{T}})
\end{align}
where $\mathbf{W}_r,\mathbf{W}_t\in \mathbb{R}^{d\times d'}$ are two linear transformations.
The readout functions here can be selected from various pooling functions (\emph{e.g.}, mean pooling). 
Our aim is for the values derived by GraphTCM to align closely with the actual correlation values.
Consequently, for the dataset $\mathcal{D}$, the set of self-superivsed tasks $\mathcal{T}=\{t_1,t_2,...,t_k\}$, the loss function for GraphTCM can be expressed as:
\begin{align}
\boldsymbol{l}=\sum_{i,j\in[1,k]}\Vert \text{GraphTCM}_{\mathcal{D}}(\mathbf{H}_{t_i}&,\mathbf{H}_{t_j})-\text{Cor}(t_i,t_j)\Vert
\end{align}
The architecture of GraphTCM is innovative and sophisticated.
GraphTCM accepts graph representations $\mathbf{H}$ as input, enabling its use even when the model architectures employed by the two self-supervised tasks differ, thus demonstrating considerable flexibility. 
Additionally, GraphTCM accounts for the non-negativity and asymmetry of correlation values, enhancing the precision and controllability of the modeling process.

\vpara{Training More Capable Representations.}
Upon considering a dataset $\mathcal{D}$ and a set of graph self-supervised tasks $\mathcal{T}=\{t_1,t_2,...,t_k\}$ along with their trained representations $\mathcal{H}=\{\mathbf{H}_{t_1},\mathbf{H}_{t_2},...,\mathbf{H}_{t_k}\}$, following the training of a GraphTCM model, we fix its parameters and utilize it to train self-supervised representations possessing high representation capability.
The entire process is notably straightforward. 
For any model $g$ with parameters $\phi$ capable of obtaining graph representations, we have:
\begin{align}
\label{formula:model_form}
\mathbf{H}_{\phi}=g_{\phi}(\mathcal{D})
\end{align}
where $g$ can represent a GNN model, an MLP, or even a linear transformation.
Our objective is to minimize the correlation values of $\mathbf{H}_{\phi}$ across various tasks.
Consequently, we aim to find the optimal parameters $\phi^*$ of the model $g$ that satisfy the following equation:
\begin{align}
\label{formula:train_high_cap}
\phi^*=\mathop{\arg\min}_{\phi}\sum_{i\in [1,k]} \frac{1}{\text{ATD}_{t_i}} \cdot \Vert \text{GraphTCM}(g_{\phi}(\mathcal{D}),\mathbf{H}_{t_i}) \Vert
\end{align}

Through the utilization of a trained GraphTCM model and employing representation capability as the objective function, we have acquired representations $\mathbf{H}_{\phi^*}$ that are expected to perform effectively across various self-supervised tasks and $\mathbf{H}_{\phi^*}$ will be applied to downstream tasks. 

\subsection{Theoretical Analysis}
\label{sec:theory}

In this chapter, we theoretically establish the relationship between representation capability and downstream task performance, demonstrating that trained representations with high representation capability lead to improved performance in downstream tasks.
For analytical simplification, we assume that the loss functions for self-supervised tasks and downstream tasks adhere to the following forms:
\begin{align}
\boldsymbol{l}=\Vert\mathbf{H}\cdot\mathbf{W}-\mathbf{Y}_{\text{target}}\Vert
\end{align}
where $\mathbf{H}$ denotes the trained representations, $\mathbf{W}$ denotes the learnable tuning head and $\mathbf{Y}_{\text{target}}$ denotes the task target.
For a set of graph self-supervised tasks $\mathcal{T}=\{t_1,t_2,...,t_k\}$ and a downstream task $t_{ds}$, we denote their $\mathbf{Y}_{\text{target}}$ as $\{\mathbf{Y}_{t_1},\mathbf{Y}_{t_2},...,\mathbf{Y}_{t_k}\}$ and $\mathbf{Y}_{t_{ds}}$.
Without loss of generality, we assume that these training targets $\mathbf{Y}_{\text{target}}$ have the same scale.
According to Definition \ref{def:cor_value}, given the trained representations $\mathcal{H}=\{\mathbf{H}_{t_1},\mathbf{H}_{t_2},...,\mathbf{H}_{t_k}\}$, the correlation values can be calculated as:
\begin{align}
\label{formula:correlation_specific_form}
\text{Cor}(t_1,t_2)=\frac{\min_{\mathbf{W}_{t_1}} \Vert\mathbf{H}_{t_1}\cdot\mathbf{W}_{t_1}-\mathbf{Y}_{t_2}\Vert}{\min_{\mathbf{W}_{t_2}}\Vert\mathbf{H}_{t_2}\cdot\mathbf{W}_{t_2}-\mathbf{Y}_{t_2}\Vert}
\end{align}
\begin{theorem}
\label{theory:loss}
Given two task $t_1,t_2\in \mathcal{T}$ and their trained representations $\mathbf{H}_{t_1},\mathbf{H}_{t_2}$, the error of $\mathbf{H}_{t_2}$ on the downstream task is $e_{t_2}$, which can be expressed as:
\begin{align}
e_{t_2}=\min_{\mathbf{W}_{t_2}^*}\Vert\mathbf{H}_{t_2}\cdot\mathbf{W}_{t_2}^*-\mathbf{Y}_{t_{ds}}\Vert 
\end{align}
Then, the error of $\mathbf{H}_{t_1}$ on the downstream task satisfies:
\begin{align}
e_{t_1}\leq \text{\rm Cor}(t_1,t_2)\cdot(e_{t_2}+\triangle)+\Vert \mathbf{Y}_{t_2}-\mathbf{Y}_{t_{ds}}\Vert
\label{formula:upper_bound}
\end{align}
where $\triangle=\Vert \mathbf{Y}_{t_2}-\mathbf{Y}_{t_{ds}}\Vert+\Vert \mathbf{H}_{t_2}\cdot(\hat{\mathbf{W}_{t_2}^*}-\hat{\mathbf{W}_{t_2}})\Vert$, and $\hat{\mathbf{W}_{t_2}^*},\hat{\mathbf{W}_{t_2}}$ denote the optimal heads for $\mathbf{W}_{t_2}^*$ and $\mathbf{W}_{t_2}$.
\end{theorem}

The complete proof of Theorem \ref{theory:loss} can be found in Appendix \ref{proof:theorem_1}. 
This theorem establishes the relationship between correlation values and downstream tasks.
According to Formula \ref{formula:upper_bound}, as $\text{Cor}(t_1,t_2)$ decreases, the upper bound on the error of representations $\mathbf{H}_{t_1}$ on downstream tasks decreases, indicating that $\mathbf{H}_{t_1}$ can achieve better downstream task performance.
In addition, we can also observe that in Formula\ref{formula:upper_bound}, there are several fixed terms related to task $t_2$, which are determined by the correlation between the self-supervised task $t_2$ and the downstream task $t_{ds}$.
However, the relationship between various self-supervised tasks and downstream tasks is unknown during the self-supervised training phase.
Therefore, we aim for trained representations to have low correlation values across all self-supervised tasks. 
If so, the representations are expected to perform well on any downstream task.
\begin{theorem}
\label{theory:new_task}
Given a set of tasks $\mathcal{T}=\{t_1,t_2,...,t_k\}$ with their trained representations $\mathcal{H}=\{\mathbf{H}_{t_1},\mathbf{H}_{t_2},...,\mathbf{H}_{t_k}\}$, for any downstream task $t_{ds}$, there exists a $\beta\geq\Vert \mathbf{Y}_{t_i}-\mathbf{Y}_{t_{ds}}\Vert+\Vert \mathbf{H}_{t_i}\cdot(\hat{\mathbf{W}_{t_i}^*}-\hat{\mathbf{W}_{t_i}})\Vert$ for any $t_i\in\mathcal{T}$.
If a new task $t'$ and its trained representations $\mathbf{H}_{t'}$ that satisfies $\delta\geq\text{\rm Cor}(t',t_i)$ for any $t_i\in\mathcal{T}$, the error of $\mathbf{H}_{t'}$ on downstream task holds:
\begin{align}
\label{formula:global}
e_{t'}\leq\delta\cdot(e_{\min}+\beta)+\beta
\end{align}
where $e_{\min}=\min\{e_{t_1},e_{t_2},...,e_{t_k}\}$.

\end{theorem}

The complete proof of Theorem \ref{theory:new_task} can be found in Appendix \ref{proof:theorem_2}. 
Formula \ref{formula:global} does not involve any specific relationship between individual self-supervised tasks and downstream tasks; instead, it provides a global perspective closer to real-world scenarios.
When we attempt to make representations have high representation capability across all self-supervised tasks, we are essentially minimizing the value of $\delta$ in Formula \ref{formula:global}, which theoretically enhances the performance of the trained representations on any downstream task.
Therefore, aiming for high representation capability and training representations according to Formula \ref{formula:train_high_cap} not only makes practical sense but also has theoretical guarantees.

\section{Experiments}
\label{sec:exp}
To systematically validate the effectiveness of our method, we answer the following three questions through experiments:
\textbf{Q1.} Can GraphTCM model the correlations between graph self-supervised tasks and demonstrate its generalization?
\textbf{Q2.} Can a well-trained GraphTCM help us obtain representations with high representation capability across all tasks?
\textbf{Q3.} Do representations trained by GraphTCM perform better on downstream tasks?
We answer these questions in Section \ref{exp:modeling}, \ref{exp:generating} and \ref{exp:evaluating} respectively.

\subsection{Experiment Setup}
\label{exp:setup}

\vpara{Datasets.} 
We employ 6 graph datasets in our experiments, including 4 citation network datasets \textit{Cora, CiteSeer, PubMed, ogbn-arxiv} and 2 product network datasets \textit{Amazon-Computers, Amazon-Photo}.
Their detailed introductions can be found in Appendix \ref{appendix:detail_dataset}.

\vpara{Involved Graph Self-supervised Tasks.}
In our experiments, we still chose the 8 graph self-supervised tasks 
(\emph{i.e.}, GraphComp, AttributeMask, GAE, EdgeMask, NodeProp, DisCluster, DGI and SubgCon) 
described in Table \ref{tab:methods} as the base tasks.
We trained 128-dimensional representations on Cora, CiteSeer, PubMed, Amazon-Computers and Amazon-Photo, and 32-dimensional representations on ogbn-arxiv.

\vpara{Implementations.}
We perform five rounds of experiments with different random seeds for each experimental setting and report the average results.
Further details of the experiments can be found in the appendix.

\subsection{Capturing the Task Correlations}
\label{exp:modeling}

\vpara{Setup.} 
We initially obtain representations trained for each self-supervised task on the dataset.
Subsequently, we calculate the correlation values between all tasks according to Definition \ref{def:cor_value}, resulting in an $8\times8$ correlation value matrix.
 This matrix serves as the training target for a GraphTCM model. During the training process, we randomly partition 70\% of the correlation values as the training set and the remaining as the validation set. Finally, we select the model with the lowest error on the validation set as the final model.
Furthermore, we introduce three new self-supervised tasks (PairAttSim~\citep{Jin2020SelfsupervisedLO}, GRACE~\citep{Zhu2020DeepGC}, and GraphMAE~\citep{Hou2022GraphMAESM}) that are not involved during training to validate the generalization capability of the trained GraphTCM. 
Specifically, we calculate the correlation values of these three tasks on the eight base tasks and compare them with the predicted values from a well-trained GraphTCM.

\begin{figure}[h]
\centering
\includegraphics[width=\columnwidth]{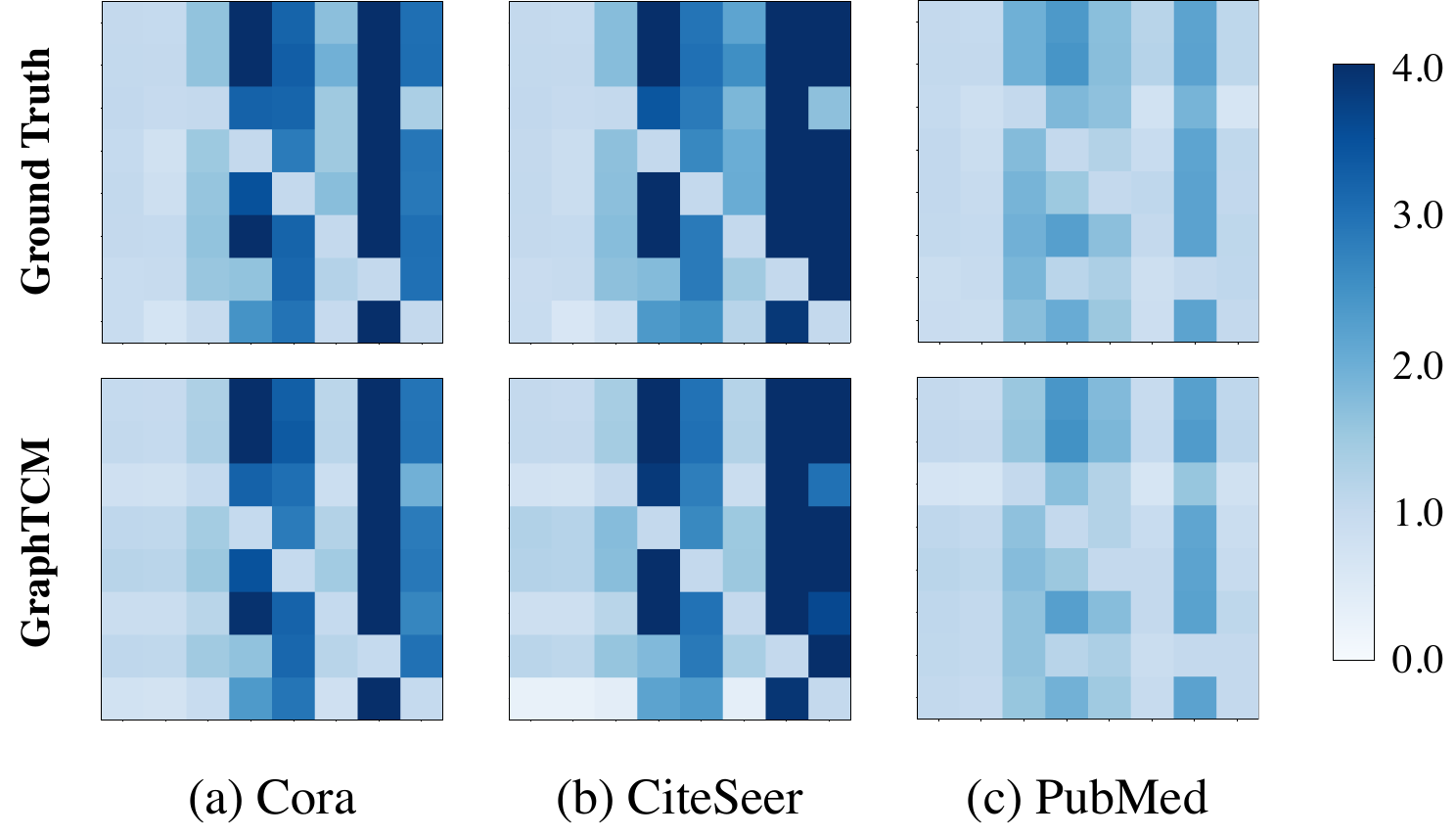}
\vspace{-0.7cm}
\caption{The modeling capability of GraphTCM. The top plots display real correlation value matrices, while the bottom plots exhibit reconstructed correlation value matrices from GraphTCM.}
\label{fig:train_result}
\vspace{-0.3cm}
\end{figure}

\begin{figure}[h]
\centering
\vspace{-0.2cm}
\includegraphics[width=\columnwidth]{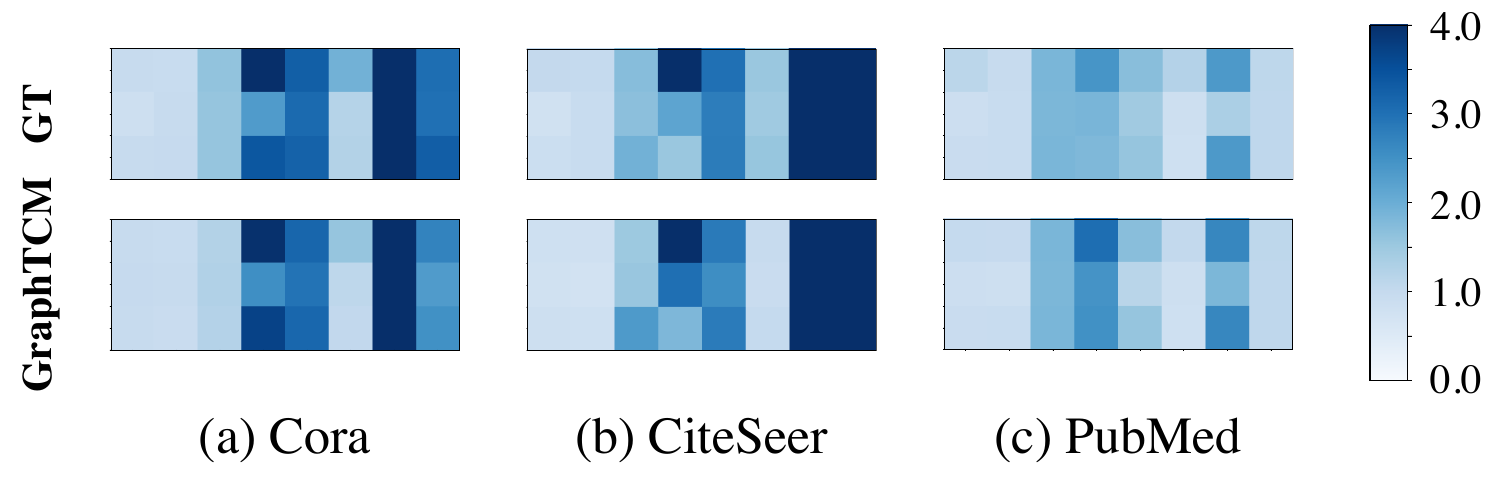}
\vspace{-0.7cm}
\caption{The generalization capability of GraphTCM. The top plots display real correlation value matrices of three unseen tasks (PairAttSim, GRACE and GraphMAE), while the bottom plots exhibit predicted correlation value matrices from GraphTCM.}
\label{fig:generalization}
\vspace{-0.4cm}
\end{figure}

\vpara{Experimental Results.}
Figure \ref{fig:train_result} illustrates the training results of GraphTCM on three datasets (Cora, CiteSeer, and PubMed).
Despite using only 70\% of the correlation values for training, GraphTCM achieves an average error of 6.04\% on the training set and 8.97\% on the validation set.
Figure \ref{fig:generalization} demonstrates the generalization results for the three new tasks.
When using a trained GraphTCM to predict the correlation values of new self-supervised tasks on the base tasks, our method achieves an average error of 13.56\%.
For the majority of graph self-supervised tasks, there already exists a 5\%-15\% random perturbation~\citep{Liu2021GraphSL}. 
Therefore, the error rate of GraphTCM is considered acceptable.

\begin{table*}[htbp]
    \centering
    \caption{The statistics of the downstream performance for trained representations of various graph self-supervised methods. The optimal results are in \textbf{bold}, while the sub-optimal results are \underline{underlined}.
    }
    \label{tab:downstream}
    \resizebox{1\textwidth}{!}{
    \begin{tabular}{c|cccccc|ccc}
    \toprule
    \multirow{2}{*}{\diagbox [width=9em, trim=lr] {\small Method}{\small Task \& Dataset}} 
    & \multicolumn{6}{c|}{Node Classification (Accuracy \%)} & \multicolumn{3}{c}{Link Prediction (ROC-AUC \%)}\\
    \cmidrule{2-7} \cmidrule{8-10}
    ~ & Cora & CiteSeer & PubMed & ogbn-arxiv & A-Computers & A-Photo & Cora & CiteSeer & PubMed\\
    \midrule
    GraphComp	& 62.04\textcolor{gray}{\small ±0.60}	& 62.74\textcolor{gray}{\small ±0.65}	& 73.92\textcolor{gray}{\small ±0.31}	& 51.10\textcolor{gray}{\small ±0.25}	& 81.81\textcolor{gray}{\small ±0.43}	& 88.32\textcolor{gray}{\small ±0.40}	& 87.54\textcolor{gray}{\small ±1.40}	& 82.50\textcolor{gray}{\small ±1.84}	& 90.95\textcolor{gray}{\small ±1.35} \\
    AttributeMask	& 59.55\textcolor{gray}{\small ±0.69}	& 59.38\textcolor{gray}{\small ±0.61}	& 73.40\textcolor{gray}{\small ±0.34}	& 53.35\textcolor{gray}{\small ±0.21}	& 81.65\textcolor{gray}{\small ±0.36}	& 88.77\textcolor{gray}{\small ±0.58}	& 81.71\textcolor{gray}{\small ±2.24}	& 80.76\textcolor{gray}{\small ±1.55}	& 88.41\textcolor{gray}{\small ±2.08} \\
    GAE	& 65.48\textcolor{gray}{\small ±0.76}	& 57.06\textcolor{gray}{\small ±0.18}	& 76.00\textcolor{gray}{\small ±0.68}	& 49.98\textcolor{gray}{\small ±0.09}	& 81.81\textcolor{gray}{\small ±0.43}	& 89.04\textcolor{gray}{\small ±0.27}	& 91.60\textcolor{gray}{\small ±0.22}	& 82.50\textcolor{gray}{\small ±1.84}	& 93.91\textcolor{gray}{\small ±0.10} \\
    EdgeMask	& 63.10\textcolor{gray}{\small ±1.01}	& 53.78\textcolor{gray}{\small ±0.40}	& 70.54\textcolor{gray}{\small ±0.67}	& 54.13\textcolor{gray}{\small ±0.24}	& 82.51\textcolor{gray}{\small ±0.28}	& 88.29\textcolor{gray}{\small ±0.40}	& \underline{92.88}\textcolor{gray}{\small ±0.24}	& 90.07\textcolor{gray}{\small ±0.64}	& 95.61\textcolor{gray}{\small ±0.09} \\
    NodeProp	& 34.10\textcolor{gray}{\small ±2.72}	& 31.58\textcolor{gray}{\small ±1.58}	& 58.22\textcolor{gray}{\small ±1.73}	& 52.14\textcolor{gray}{\small ±0.15}	& 81.08\textcolor{gray}{\small ±0.38}	& 86.77\textcolor{gray}{\small ±0.35}	& 83.31\textcolor{gray}{\small ±1.80}	& 84.49\textcolor{gray}{\small ±0.91}	& 93.12\textcolor{gray}{\small ±0.67} \\
    DisCluster	& 64.97\textcolor{gray}{\small ±1.40}	& 62.18\textcolor{gray}{\small ±1.12}	& 70.74\textcolor{gray}{\small ±0.43}	& 53.33\textcolor{gray}{\small ±0.15}	& 81.62\textcolor{gray}{\small ±0.61}	& 88.65\textcolor{gray}{\small ±0.33}	& 84.71\textcolor{gray}{\small ±1.84}	& 86.02\textcolor{gray}{\small ±2.25}	& 91.93\textcolor{gray}{\small ±1.04} \\
    DGI	& 76.92\textcolor{gray}{\small ±0.42}	& 67.01\textcolor{gray}{\small ±0.38}	& \underline{76.96}\textcolor{gray}{\small ±0.24}	& 50.97\textcolor{gray}{\small ±0.13}	& 81.66\textcolor{gray}{\small ±0.50}	& 88.53\textcolor{gray}{\small ±0.50}	& 92.04\textcolor{gray}{\small ±2.07}	& 89.46\textcolor{gray}{\small ±1.30}	& 95.00\textcolor{gray}{\small ±0.04} \\
    SubgCon	& \underline{77.28}\textcolor{gray}{\small ±0.15}	& 68.57\textcolor{gray}{\small ±0.51}	& 76.76\textcolor{gray}{\small ±0.56}	& 52.56\textcolor{gray}{\small ±0.30}	& 80.25\textcolor{gray}{\small ±0.31}	& 87.33\textcolor{gray}{\small ±0.08}	& 91.29\textcolor{gray}{\small ±0.17}	& 91.23\textcolor{gray}{\small ±0.29}	& 91.29\textcolor{gray}{\small ±0.14} \\
    \midrule
    Addition	& 74.12\textcolor{gray}{\small ±0.47}	& 67.86\textcolor{gray}{\small ±0.48}	& 75.64\textcolor{gray}{\small ±0.93}	& 52.16\textcolor{gray}{\small ±0.24}	& 83.57\textcolor{gray}{\small ±0.23}	& 90.02\textcolor{gray}{\small ±0.24}	& 92.54\textcolor{gray}{\small ±0.13}	& 92.29\textcolor{gray}{\small ±0.29}	& 94.92\textcolor{gray}{\small ±0.14} \\
    Concat	& 76.30\textcolor{gray}{\small ±0.37}	& 67.92\textcolor{gray}{\small ±0.54}	& 76.74\textcolor{gray}{\small ±0.22}	& \underline{53.88}\textcolor{gray}{\small ±0.44}	& 82.43\textcolor{gray}{\small ±0.38}	& \underline{91.07}\textcolor{gray}{\small ±0.42}	& 92.86\textcolor{gray}{\small ±0.04}	& \underline{92.59}\textcolor{gray}{\small ±0.16}	& 94.44\textcolor{gray}{\small ±0.10} \\
    \midrule
    Multi-loss	& 57.76\textcolor{gray}{\small ±2.87}	& 52.74\textcolor{gray}{\small ±3.10}	& 67.08\textcolor{gray}{\small ±2.09}	& 51.26\textcolor{gray}{\small ±0.13}	& 81.55\textcolor{gray}{\small ±0.72}	& 90.06\textcolor{gray}{\small ±0.38}	& 85.70\textcolor{gray}{\small ±1.17}	& 83.88\textcolor{gray}{\small ±2.83}	& 92.52\textcolor{gray}{\small ±0.79} \\
    AutoSSL	& 76.64\textcolor{gray}{\small ±0.71}	& \underline{70.30}\textcolor{gray}{\small ±0.46}	& 75.78\textcolor{gray}{\small ±0.69}	& 53.45\textcolor{gray}{\small ±0.23}	& \underline{83.42}\textcolor{gray}{\small ±0.30}	& 90.70\textcolor{gray}{\small ±0.41}	& 92.00\textcolor{gray}{\small ±3.66}	& 92.01\textcolor{gray}{\small ±0.04}	& \underline{96.55}\textcolor{gray}{\small ±0.36} \\
    ParetoGNN	& 52.03\textcolor{gray}{\small ±0.30}	& 59.73\textcolor{gray}{\small ±0.50}	& 76.54\textcolor{gray}{\small ±0.50}	& 53.84\textcolor{gray}{\small ±0.26}	& 79.26\textcolor{gray}{\small ±0.26}	& 84.50\textcolor{gray}{\small ±0.53}	& 77.29\textcolor{gray}{\small ±5.20}	& 87.09\textcolor{gray}{\small ±4.65}	& 96.06\textcolor{gray}{\small ±0.18} \\
    \midrule
    GraphTCM (Ours)	& \textbf{81.50}\textcolor{gray}{\small ±0.52}	& \textbf{72.83}\textcolor{gray}{\small ±0.64}	& \textbf{77.24}\textcolor{gray}{\small ±0.48}	& \textbf{54.73}\textcolor{gray}{\small ±0.18}	& \textbf{84.95}\textcolor{gray}{\small ±0.34}	& \textbf{92.06}\textcolor{gray}{\small ±0.19}	& \textbf{95.52}\textcolor{gray}{\small ±0.15}	& \textbf{96.14}\textcolor{gray}{\small ±0.24}	& \textbf{97.13}\textcolor{gray}{\small ±0.06} \\
    \bottomrule
    \end{tabular}
    }
    \vspace{-0.3cm}
\end{table*}

\subsection{Generating Capable Representations}
\label{exp:generating}

\vpara{Setup.}
Once we have trained a GraphTCM on a particular dataset, we can leverage it to train representations with high representation capability.
According to Formula \ref{formula:model_form} and \ref{formula:train_high_cap}, GraphTCM can help to train models of arbitrary forms.
In our experiments, we only train a linear transformation to obtain the GraphTCM-enhanced representations.
Specifically, given a set of tasks $\mathcal{T}=\{t_1,t_2,...,t_k\}$ with their trained representations $\mathcal{H}=\{\mathbf{H}_{t_1},\mathbf{H}_{t_2},...,\mathbf{H}_{t_k}\}\in\mathbb{R}^{N\times d}$, we train a matrix $\mathbf{W}\in\mathbb{R}^{k\times d}$ and our GraphTCM-enhanced representations $\mathbf{H}'$ satisfy $\mathbf{H}'(a,b)=\sum_i^k \mathbf{H}_{t_i}(a,b)\cdot\mathbf{W}(i,b)$, where $\mathbf{H}(a,b)$ denotes the value at the $a$-th row and the $b$-th column in $\mathbf{H}$. 
During the training process, we optimize the parameters of $\mathbf{W}$ according to Formula \ref{formula:train_high_cap}.

\begin{figure}[h]
\centering
\vspace{-0.3cm}
\includegraphics[width=\columnwidth]{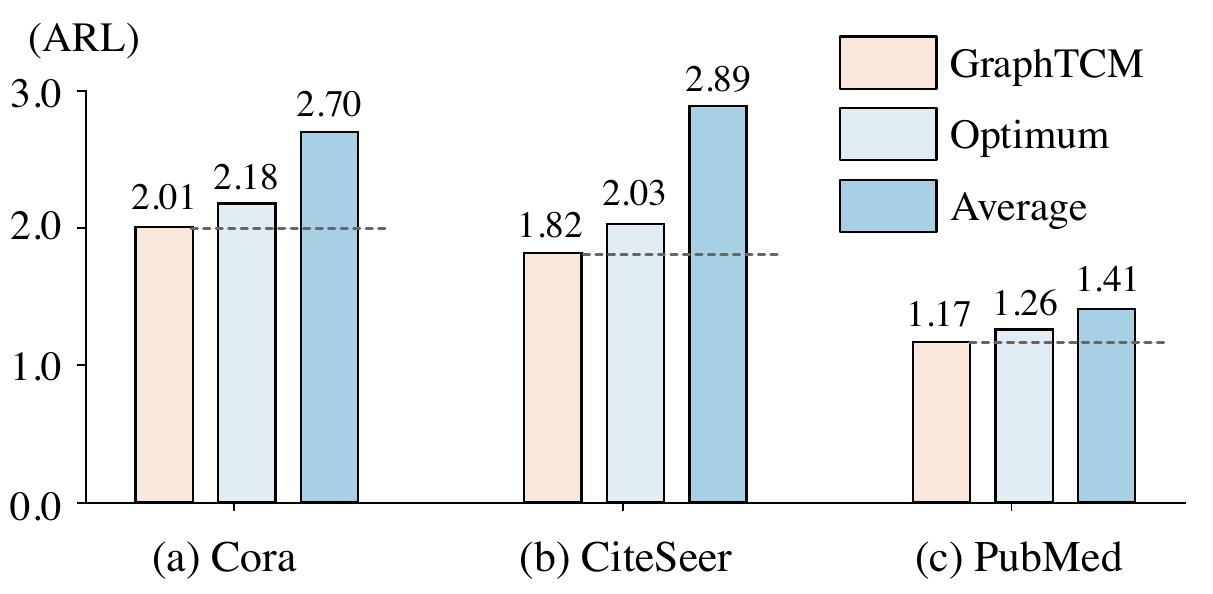}
\vspace{-0.8cm}
\caption{The statistics of ARL values for representations trained by GraphTCM and base tasks (``Optimum'' represents the lowest ARL values among base tasks while ``Average'' represents the average ARL values for base tasks).}
\label{fig:train_cap}
\vspace{-0.8cm}
\end{figure}

\vpara{Experimental Results.}
Figure \ref{fig:train_cap} illustrates the comparison of ARL values for our method and base tasks.
Our GraphTCM-enhanced representations achieve the lowest ARL values across all datasets.
Specifically, on three datasets (Cora, CiteSeer and PubMed), the ARL values of the GraphTCM-enhanced representations decrease by an average of 8.42\% compared to the lowest ARL values of the base tasks and decrease by an average of 26.53\% compared to the average ARL values of the base tasks.
Our experimental results demonstrate the effectiveness of GraphTCM, indicating that we can obtain representations with high capability with the assistance of a trained GraphTCM.

\subsection{Evaluating on Downstream Tasks}
\label{exp:evaluating}

\vpara{Setup.}
Next, we evaluate the performance of the trained representations on downstream tasks.
During the downstream training, we fix the representations obtained from the self-supervised tasks and only train linear tuning heads to adapt the self-supervised representations to the downstream tasks.
We conduct self-supervised training on the six datasets described in Section \ref{exp:setup} and evaluate the performance of the trained representations for downstream node classification on these datasets.
In addition, we select three datasets (Cora, CiteSeer and PubMed) to validate the ability of the self-supervised representations for link prediction.
To further demonstrate the effectiveness of GraphTCM, we involve three multi-task methods (Multi-loss, AutoSSL, ParetoGNN) and two naive representation mixing methods (representation addition and concat) as our baselines.
Their detailed descriptions can be found in Appendix \ref{appendix:detail_multi_task}.

\vpara{Experimental Results.}
Table \ref{tab:downstream} presents the comprehensive results for various graph self-supervised tasks. Our GraphTCM-enhanced representations achieve optimal performance across \textit{all} datasets and downstream tasks.
For node classification tasks, our GraphTCM-enhanced representations exhibit an average improvement of 1.7\% compared to the sub-optimal self-supervised representations. 
Meanwhile, for link prediction, the average improvement is 2.2\%.
However, the performance of representations trained by multi-task methods falls far short of expectations. 
In most cases, these methods are inferior to concatenating trained representations from base tasks. 
Moreover, in some situations, the representations trained by these multi-task methods even exhibit significant performance degradation compared to the representations trained by single base tasks. 
In summary, our proposed GraphTCM successfully models the correlations between self-supervised tasks and can be used to train representations with high capability. 
On the downstream tasks, GraphTCM-enhanced representations demonstrate far superior effectiveness and universality compared to representations trained by other methods.

\section{Conclusion}

In this paper, we have unveiled the task correlations for graph self-supervised learning by calculating the correlation values between various graph self-supervised tasks.
By demonstrating the actual correlation values between tasks across various datasets, we reveal the complexity of task correlations and the limitations of existing multi-task learning methods. 
To obtain more capable representations, we propose GraphTCM to model these task correlations. 
The experimental results indicate that GraphTCM can effectively characterize task correlations, and GraphTCM-enhanced representations have achieved outstanding performance on various downstream tasks.

\newpage
\section*{Impact Statement}
This paper presents work whose goal is to advance the field of Machine Learning. There are many potential societal consequences of our work, none which we feel must be specifically highlighted here.
\section*{Acknowledgements}
This work is supported by National Natural Science Foundation of China (No. 62322606, No. 62441605).

\bibliography{reference}
\bibliographystyle{icml2024}

\newpage
\appendix
\onecolumn
\section{Extra Materials for Section \ref{sec:method}}

\subsection{The Detailed Descriptions of Involved Self-supervised Tasks}
\label{appendix:detailed_method}

In our pursuit of a comprehensive understanding of the intricate relationships within graph self-supervised tasks, we have carefully selected two exemplary self-supervised tasks for each category. Detailed introductions to these tasks are provided as below.
\begin{itemize}
    \item {\textit{GraphComp}~\citep{You2020WhenDS}. Analogous to image completion, GraphComp involves masking the features of specific target nodes, which are then input into the Graph Convolutional Networks (GCNs). The objective is to reconstruct the masked features, teaching the network to extract features from the context, thereby classifying it as a typical feature-based (FB)} category. 
    \item {\textit{AttributeMask}~\citep{Jin2020SelfsupervisedLO}. AttributeMask aims to reconstruct the dense feature matrix generated by Principal Component Analysis (PCA) rather than the raw features, addressing challenges associated with reconstructing high-dimensional and sparse feature representations.} 
    \item {\textit{GAE}~\citep{Kipf2016VariationalGA}. The graph autoencoder (GAE) is a straightforward example of a structure generation method. Within the GAE framework, a GCN encoder initially generates node representations from the original graph. Subsequently, an inner production function with a sigmoid activation function acts as the decoder, aiming to reconstruct the adjacency matrix using the node representations.}
    \item {\textit{EdgeMask}~\citep{Jin2020SelfsupervisedLO}. The EdgeMask task aims to acquire finer-grained local structural information by employing link prediction as a pretext task. It selectively masks certain edge connections, and the primary learning objective revolves around predicting the presence or absence of a link between a specified node pair.} 
    \item {\textit{NodeProp}~\citep{Jin2020SelfsupervisedLO}. NodeProp utilizes a node-level pretext task, predicting properties for individual nodes, including attributes such as degree, local node importance, and local clustering coefficient.}
    \item {\textit{DisCluster}~\citep{Jin2020SelfsupervisedLO}. DisCluster’s objective is to perform regression on the distances between each node and predefined graph clusters. It partitions the graph into multiple clusters and designates the node with the highest degree within each cluster as its cluster center. The task is to predict the distances from each node to all cluster centers.}
    \item {\textit{DGI}~\citep{Velickovic2018DeepGI}. DGI maximizes mutual information between representations from subgraphs with differing scales, facilitating the graph encoder in attaining a comprehensive grasp of both localized and global semantic information.}
    \item {\textit{SubgCon}~\citep{jiao2020sub}. Sub-graph Contrast (SubgCon) captures regional structural insights by capitalizing on the robust correlation between central nodes and their sampled subgraphs. It fosters the learning of node representations via a contrastive loss defined using subgraphs sampled from the original graph, adopting an innovative data augmentation strategy.}
\end{itemize}

The training objectives of the self-supervised methods we have chosen are well-defined and singular, widely applicable across diverse graph self-supervised scenarios.
These methods are highly representative, and consequently, the majority of existing graph self-supervised methods can be broken down into combinations of the aforementioned self-supervised tasks.

\subsection{The Detailed Descriptions of Involved Multi-task Training Methods}
\label{appendix:detail_multi_task}

Given a set of base graph self-supervised tasks $\mathcal{T}=\{t_1,t_2,...,t_k\}$ with their trained representations $\mathcal{H}=\{\mathbf{H}_{t_1},\mathbf{H}_{t_2},...,\mathbf{H}_{t_k}\}$ and losses $\mathcal{L}=\{\boldsymbol{l}_{t_1},\boldsymbol{l}_{t_2},...,\boldsymbol{l}_{t_k}\}$, these methods can be expressed as below:
\begin{itemize}
    \item {\textit{Addition, Concat.}  Representation addition sums all the trained representations to obtain $\mathbf{H}_{\text{add}}=\sum_i^{k}\mathbf{H}_{t_i}$ while representation concat concatenates all the trained representations to obtain $\mathbf{H}_{\text{concat}}=[\mathbf{H}_{t_1}|\mathbf{H}_{t_2}|...|\mathbf{H}_{t_k}]$. 
    }
    \item {\textit{Multi-loss.} These methods combine the losses from multiple tasks for self-supervised training. Multi-loss combines the losses from different tasks with weighted summation $\boldsymbol{l}=\sum\limits_i^k \alpha_i \boldsymbol{l}_{t_i}$, where $\alpha_1,\alpha_2,...,\alpha_k$ are learnable task weight coefficients. }
    \item {\textit{AutoSSL.} AutoSSL~\citep{Jin2021AutomatedSL} can automatically search over combinations of various self-supervised tasks using their pseudo-homophily. Specifically, its optimization objective is defined as $\mathop {\min }\limits_{{\lambda _1},...,{\lambda _k}} \mathcal{H}(\theta ^*),s.t.\theta ^* = \arg \mathop {\min }\limits_\theta  \mathcal{L}(\{\theta_i \},\{ {\alpha _i}\} ,\{ {l_i}\} ) = \arg \mathop {\min }\limits_{\theta_i} \sum\limits_{i = 1}^k {{\alpha _i}{l_i}} $, where H represents an unsupervised quality metric employed to assess the quality of the acquired embedding. $\alpha_i$ ,$\lambda_i$ and $\theta$ correspond to the underlying self-supervised task ,their respective weights and the learnable parameters of neural networks.}
    \item {\textit{ParetoGNN.} ParetoGNN~\citep{Ju2022MultitaskSG} dynamically allocates task weights to facilitate Pareto optimality. Specifically, within the combination of self-supervised tasks, it is formulated as $\boldsymbol{l}=\sum\limits_i^k \alpha_i \boldsymbol{l}_{t_i}$, where $\alpha_i$ represents the task weight coefficient.}
\end{itemize}

\subsection{What Makes Better Representations}
Under ideal conditions, our objective is to acquire representations through self-supervised training that demonstrate low correlation values across all self-supervised tasks. 
Naturally, representations trained with minimal correlation across these tasks should possess the expressive capacity necessary for all self-supervised tasks. 
Typically, such trained representations will also display strong generality and deliver high performance in downstream tasks. 
Therefore, it is entirely reasonable for us to establish high representation capability as our training goal. 
In Section \ref{sec:theory}, we theoretically analyze the relationship between representation capability and downstream performance.
Our theoretical findings align with our intuition, indicating that high representation capability significantly contributes to enhanced performance in downstream tasks.
Thus, in comparison to task difficulty, our primary focus lies in the representation capability of the task, as it directly impacts the performance of the trained representation in downstream tasks.

\subsection{Task Correlations Within the Same Category and Across Different Categories}
Intuitively, graph self-supervised tasks within the same category should exhibit stronger mutual correlations compared to tasks from different categories. However, our experiments do not consistently support this notion; only structure-based graph self-supervised tasks demonstrate such a tendency. Feature-based tasks, on the other hand, tend to have relatively lower task difficulty, indicating that other self-supervised tasks have smaller correlation values with feature-based tasks. Consequently, the correlation between different feature-based tasks is not notably strong. Similarly, auxiliary property-based tasks often show significant variations in their practical implementations, with no explicit connection between tasks such as node degree prediction and node clustering. Hence, different auxiliary property-based tasks also do not demonstrate stronger task correlations between them. The same observation applies to contrast-based tasks, where different tasks have notably different objectives. Conversely, structure-based tasks typically center on graph structure reconstruction, leading to stronger correlations among tasks in this category. In general, self-supervised tasks with closely aligned objectives should exhibit stronger task correlations. However, it is crucial to note that not all tasks within the same category have closely aligned objectives.

\subsection{Proof of Theorem \ref{theory:loss}}
\label{proof:theorem_1}
\textbf{Theorem 3.4.} \textit{Given two task $t_1,t_2\in \mathcal{T}$ and their trained representations $\mathbf{H}_{t_1},\mathbf{H}_{t_2}$, the error of $\mathbf{H}_{t_2}$ on the downstream task is $e_{t_2}$, which can be expressed as:
\begin{align}
e_{t_2}=\min_{\mathbf{W}_{t_2}^*}\Vert\mathbf{H}_{t_2}\cdot\mathbf{W}_{t_2}^*-\mathbf{Y}_{t_{ds}}\Vert \nonumber
\end{align}
Then, the error of $\mathbf{H}_{t_1}$ on the downstream task satisfies:
\begin{align}
e_{t_1}\leq \text{\rm Cor}(t_1,t_2)\cdot(e_{t_2}+\triangle)+\Vert \mathbf{Y}_{t_2}-\mathbf{Y}_{t_{ds}}\Vert \nonumber
\end{align}
where $\triangle=\Vert \mathbf{Y}_{t_2}-\mathbf{Y}_{t_{ds}}\Vert+\Vert \mathbf{H}_{t_2}\cdot(\hat{\mathbf{W}_{t_2}^*}-\hat{\mathbf{W}_{t_2}})\Vert$, and $\hat{\mathbf{W}_{t_2}^*},\hat{\mathbf{W}_{t_2}}$ denote the optimal heads for $\mathbf{W}_{t_2}^*$ and $\mathbf{W}_{t_2}$.}

\begin{proof}

According to Formula \ref{formula:correlation_specific_form}, the correlation value $\text{Cor}(t_1,t_2)$ can be expressed as:
\begin{align}
\text{Cor}(t_1,t_2)=\frac{\min_{\mathbf{W}_{t_1}} \Vert\mathbf{H}_{t_1}\cdot\mathbf{W}_{t_1}-\mathbf{Y}_{t_2}\Vert}{\min_{\mathbf{W}_{t_2}}\Vert\mathbf{H}_{t_2}\cdot\mathbf{W}_{t_2}-\mathbf{Y}_{t_2}\Vert}
\end{align}
We assume that for task $t_2$, the optimal heads for $\mathbf{H}_{t_1}$ and $\mathbf{H}_{t_2}$ are $\hat{\mathbf{W}}_{t_1}$ and $\hat{\mathbf{W}}_{t_2}$.
Meanwhile, the optimal head for $\mathbf{H}_{t_2}$ on downstream task is $\hat{\mathbf{W}_{t_2}^*}$:
\begin{align}
\hat{\mathbf{W}}_{t_1}=\arg\min_{\mathbf{W}_{t_1}} \Vert\mathbf{H}_{t_1}\cdot\mathbf{W}_{t_1}-\mathbf{Y}_{t_2}\Vert \\
\hat{\mathbf{W}}_{t_2}=\arg\min_{\mathbf{W}_{t_2}} \Vert\mathbf{H}_{t_2}\cdot\mathbf{W}_{t_2}-\mathbf{Y}_{t_2}\Vert \\
\hat{\mathbf{W}}_{t_2}^*=\arg\min_{\mathbf{W}_{t_2}^*} \Vert\mathbf{H}_{t_2}\cdot\mathbf{W}_{t_2}^*-\mathbf{Y}_{t_{ds}}\Vert
\end{align}
Then, the error of $\mathbf{H}_{t_1}$ on the downstream tasks can be calculated as follows:
\begin{align}
e_{t_1}&=\min_{\mathbf{W}_{t_1}^*}\Vert\mathbf{H}_{t_1}\cdot\mathbf{W}_{t_1}^*-\mathbf{Y}_{t_{ds}}\Vert \\
&=\min_{\mathbf{W}_{t_1}^*}\Vert\mathbf{H}_{t_1}\cdot\mathbf{W}_{t_1}^*-\mathbf{Y}_{t_{2}}+\mathbf{Y}_{t_{2}}-\mathbf{Y}_{t_{ds}}\Vert \\
&\leq \min_{\mathbf{W}_{t_1}^*}\Vert\mathbf{H}_{t_1}\cdot\mathbf{W}_{t_1}^*-\mathbf{Y}_{t_{2}}\Vert+\Vert\mathbf{Y}_{t_{2}}-\mathbf{Y}_{t_{ds}}\Vert \\
&=\Vert \mathbf{H}_{t_1}\cdot\hat{\mathbf{W}}_{t_1}-\mathbf{Y}_{t_{2}} \Vert + \Vert\mathbf{Y}_{t_{2}}-\mathbf{Y}_{t_{ds}}\Vert \\
&=\text{Cor}(t_1,t_2)\cdot\Vert \mathbf{H}_{t_2}\cdot\hat{\mathbf{W}}_{t_2}-\mathbf{Y}_{t_{2}}\Vert + \Vert\mathbf{Y}_{t_{2}}-\mathbf{Y}_{t_{ds}}\Vert \\
&=\text{Cor}(t_1,t_2)\cdot\Vert \mathbf{H}_{t_2}\cdot\hat{\mathbf{W}}_{t_2}^*-\mathbf{Y}_{t_{ds}}+\mathbf{H}_{t_2}\cdot(\hat{\mathbf{W}}_{t_2}-\hat{\mathbf{W}}_{t_2}^*)+ \mathbf{Y}_{t_{ds}}-\mathbf{Y}_{t_2}\Vert + \Vert\mathbf{Y}_{t_{2}}-\mathbf{Y}_{t_{ds}}\Vert \\
&\leq \text{Cor}(t_1,t_2)\cdot(\Vert \mathbf{H}_{t_2}\cdot\hat{\mathbf{W}}_{t_2}^*-\mathbf{Y}_{t_{ds}}\Vert+\Vert\mathbf{H}_{t_2}\cdot(\hat{\mathbf{W}}_{t_2}-\hat{\mathbf{W}}_{t_2}^*)\Vert+ \Vert\mathbf{Y}_{t_{ds}}-\mathbf{Y}_{t_2}\Vert) + \Vert\mathbf{Y}_{t_{2}}-\mathbf{Y}_{t_{ds}}\Vert\\
&=\text{Cor}(t_1,t_2)\cdot(e_{t_2}+\Vert\mathbf{H}_{t_2}\cdot(\hat{\mathbf{W}}_{t_2}-\hat{\mathbf{W}}_{t_2}^*)\Vert+ \Vert\mathbf{Y}_{t_{ds}}-\mathbf{Y}_{t_2}\Vert)+\Vert\mathbf{Y}_{t_{2}}-\mathbf{Y}_{t_{ds}}\Vert\label{formula:appendix_theory_one_final}
\end{align}
If we denote $\triangle=\Vert \mathbf{Y}_{t_2}-\mathbf{Y}_{t_{ds}}\Vert+\Vert \mathbf{H}_{t_2}\cdot(\hat{\mathbf{W}_{t_2}^*}-\hat{\mathbf{W}_{t_2}})\Vert$, then we have:
\begin{align}
e_{t_1}\leq\text{Cor}(t_1,t_2)\cdot(e_{t_2}+\triangle)+\Vert\mathbf{Y}_{t_{2}}-\mathbf{Y}_{t_{ds}}\Vert
\end{align}
Thus, we have proved the correctness of Theorem \ref{theory:loss}.

\end{proof}

\subsection{Proof of Theorem \ref{theory:new_task}}
\label{proof:theorem_2}
\textbf{Theorem 3.5.} \textit{Given a set of tasks $\mathcal{T}=\{t_1,t_2,...,t_k\}$ with their trained representations $\mathcal{H}=\{\mathbf{H}_{t_1},\mathbf{H}_{t_2},...,\mathbf{H}_{t_k}\}$, for any downstream task $t_{ds}$, there exists a $\beta\geq\Vert \mathbf{Y}_{t_i}-\mathbf{Y}_{t_{ds}}\Vert+\Vert \mathbf{H}_{t_i}\cdot(\hat{\mathbf{W}_{t_i}^*}-\hat{\mathbf{W}_{t_i}})\Vert$ for any $t_i\in\mathcal{T}$.
If a new task $t'$ and its trained representations $\mathbf{H}_{t'}$ that satisfies $\delta\geq\text{\rm Cor}(t',t_i)$ for any $t_i\in\mathcal{T}$, the error of $\mathbf{H}_{t'}$ on downstream task holds:
\begin{align}
e_{t'}\leq\delta\cdot(e_{\min}+\beta)+\beta \nonumber
\end{align}
where $e_{\min}=\min\{e_{t_1},e_{t_2},...,e_{t_k}\}$.}

\begin{proof}

According to Formula \ref{formula:appendix_theory_one_final}, for any task $t_i\in\mathcal{T}$, we can obtain the following inequality:
\begin{align}
e_{t'}\leq\text{Cor}(t',t_i)\cdot(e_{t_i}+\Vert\mathbf{H}_{t_i}\cdot(\hat{\mathbf{W}}_{t_i}-\hat{\mathbf{W}}_{t_i}^*)\Vert+ \Vert\mathbf{Y}_{t_{ds}}-\mathbf{Y}_{t_i}\Vert)+\Vert\mathbf{Y}_{t_i}-\mathbf{Y}_{t_{ds}}\Vert \label{formula:appendix_theory_2}
\end{align}
We have $\forall t_i\in\mathcal{T},\beta\geq\Vert \mathbf{Y}_{t_i}-\mathbf{Y}_{t_{ds}}\Vert+\Vert \mathbf{H}_{t_i}\cdot(\hat{\mathbf{W}_{t_i}^*}-\hat{\mathbf{W}_{t_i}})\Vert $.
Formula \ref{formula:appendix_theory_2} can be transformed into the following form:
\begin{align}
\forall t_i\in\mathcal{T}\quad e_{t'}\leq\text{Cor}(t',t_i)\cdot(e_{t_i}+\beta)+\beta
\end{align}
We also have $\forall t_i\in\mathcal{T},\delta\geq\text{\rm Cor}(t',t_i)$.
Then, we can obtain:
\begin{align}
\forall t_i\in\mathcal{T}\quad e_{t'}\leq\delta\cdot(e_{t_i}+\beta)+\beta
\end{align}
Therefore, we can obtain the conclusion in Theorem \ref{theory:new_task}:
\begin{align}
e_{t'}\leq\delta\cdot(e_{\min}+\beta)+\beta 
\end{align}
where $e_{\min}=\min\{e_{t_1},e_{t_2},...,e_{t_k}\}$.

\end{proof}

\section{More Information on Experiments}

\subsection{Detailed Descriptions of Datasets}
\label{appendix:detail_dataset}
We employ 6 graph datasets in our experiments, including 4 citation network datasets \textit{Cora, CiteSeer, PubMed, ogbn-arxiv} and 2 product network datasets \textit{Amazon-Computers, Amazon-Photo}.
Table \ref{tab:statistics} presents the statistics of involved datasets.
\begin{itemize}
    \item {\textit{Cora, CiteSeer, PubMed, ogbn-arxiv:}
    These 4 different citation networks are widely used as graph benchmarks~\citep{sen2008collective,hu2020ogb}. Their downstream task is to determine the research area of papers/researchers.}
    \item {\textit{Amazon-Computers, Amazon-Photo:}
    These are 2 product networks~\citep{Shchur2018PitfallsOG} from Amazon. The nodes here represent goods and edges represent that two goods are frequently bought together. The target is to map goods to their respective product category.}
\end{itemize}

\begin{table*}[h]
	\centering
	\caption{The statistics of involved datasets.}
	\label{tab:statistics}
	\resizebox{0.7\textwidth}{!}{
	\begin{tabular}{c|ccccc}
	\toprule
	Dataset	& Nodes	& Edges	& Feature & Classes & Train/Val/Test\\
	\midrule
	Cora & 2708 & 5429 & 1433 & 7 & 140 / 500 / 1000\\
    CiteSeer & 3327 & 4732 & 3703 & 6 & 120 / 500 / 1000\\ 
	PubMed & 19717 & 44338 & 500 & 3 & 60 / 500 / 1000\\
	ogbn-arxiv & 169343 & 1166243 & 128 & 40 & 90941 / 29799 / 48603\\ 
	Amazon-Computers & 13752 & 491722 & 767 & 10 & 10\% / 10\% / 80\%\\
	Amazon-Photo & 7650 & 238162 & 745 & 8 & 10\% / 10\% / 80\%\\
	\bottomrule
    \end{tabular}
    }
\end{table*}

\subsection{Training the Self-supervised Representations}

The hyperparameters for the aforementioned Self-supervised Tasks can be found in Table~\ref{sslhyper}. 
All backbone models for these tasks have utilized the Graph Convolutional Network (GCN) with a linear architecture, and the parameter settings are largely consistent with the original paper. 
We trained 128-dimensional representations on Cora, CiteSeer, PubMed, Amazon-Computers, and Amazon-Photo, and 32-dimensional representations on ogbn-arxiv.
Additionally, for the GraphComp model, the (learning rate, weight decay) values are set to (5e-4, 0.7) for the \textit{Citeseer} dataset and (5e-4, 0.5) for the \textit{Pubmed} dataset.

\begin{table*}[h]
\centering
\caption{The hyperparameters of training self-supervised representations. ES indicates the adoption of the early stopping strategy; \# GCNs and \# linears respectively denote the number of GCNs and linear layers in the self-supervised task.}
\label{sslhyper}
\resizebox{0.8\textwidth}{!}
{
\begin{tabular}{c|ccccccccc}
\toprule
& optimizer & learning rate & weight decay & dropout & epochs & \# GCNs & \# linears \\ 
\midrule
GraphComp     & Adam & 0.008 & 8e-5 & 0.5 &  500  & 0 & 3 \\
AttributeMask & Adam & 0.001 & 5e-4 & 0.5 &  200  & 1 & 1  \\
\midrule
GAE           & Adam & 0.01  &  0   &  0  &  500  & 2 & 0 \\
EdgeMask      & Adam & 0.001 & 5e-4 & 0.5 &  200  & 1 & 1  \\
\midrule
NodeProp      & Adam & 0.001 & 5e-4 & 0.5 &  200  & 1 & 1  \\
DisCluster    & Adam & 0.001 & 5e-4 & 0.5 &  200  & 1 & 1  \\
\midrule
DGI           & Adam & 0.001 &  0   &  0  &  ES & 1 & 1  \\
SubgCon       & Adam & 0.001 &  0   &  0  &  50  & 1 & 0  \\
\bottomrule
\end{tabular}
}
\end{table*}

\subsection{Adapting to Downstream Tasks}

When applying the trained representations to downstream tasks, we exclusively train a linear transformation $\mathbf{W}\in \mathbb{R}^{d\times d'}$, where $d$ denotes the dimension of the trained representations, and $d'$ denotes the dimension required for the downstream tasks.
Across all datasets and methods, we maintain consistent hyperparameters, setting the learning rate to 0.001, weight decay to 0.0005, and conducting 300 training epochs.

\subsection{Ablation Study}
GraphTCM aims to estimate correlation values between various graph self-supervised tasks. To validate the effectiveness of each component, we designed three variants: GraphTCM w/o $\mathbf{W}_r$, GraphTCM w/o $\mathbf{W}_t$, and GraphTCM w/o Exp. Specifically, GraphTCM w/o $\mathbf{W}_r$ removes the learnable parameter $\mathbf{W}_r$, GraphTCM w/o $\mathbf{W}_t$ removes the learnable parameter$\mathbf{W}_t$, while GraphTCM w/o Exp removes the exponential function used in computing correlation values. Table \ref{tab:ablation} presents the average relative errors of each variant in characterizing the correlation values on three citation datasets: Cora, CiteSeer and PubMed.

\begin{table*}[h]
	\centering
	\caption{The average relative errors of each variant.}
	\label{tab:ablation}
	\resizebox{0.7\textwidth}{!}{
	\begin{tabular}{c|cc}
	\toprule
	Model & Average training error (\%) & Average validating error (\%)\\
	\midrule
        GraphTCM & 6.04 & 8.97\\
        GraphTCM w/o $\mathbf{W}_r$ & 83.44 & 105.36\\
        GraphTCM w/o $\mathbf{W}_t$ & 91.76 & 128.30\\
        GraphTCM w/o Exp & 21.03 & 51.05\\
	\bottomrule
    \end{tabular}
    }
\end{table*}

\end{document}